\documentclass[11pt]{article}

\usepackage[margin=1in]{geometry}
\usepackage{booktabs}
\usepackage{amsmath}
\usepackage{graphicx}
\graphicspath{{figures/}}
\usepackage{listings}
\usepackage{xcolor}
\usepackage[hidelinks]{hyperref}
\usepackage{microtype}

\title{What Irregularity Costs:\\
CUDA C++, Rust, and Triton on a Hash-Blocked GPU Workload}

\author{Petr Korolev \\ Spacial Intelligence Labs}
\date{\today}

\begin{document}
\maketitle

\begin{abstract}
GPU language comparisons are almost always run on tiled dense linear algebra, where every toolchain is good and the differences are small. We implement the same hash-blocked TSDF fusion kernel in CUDA C++, in Rust through NVIDIA's \texttt{cuda-oxide}, and in Triton, and measure it on a workload with the opposite character: an open-addressed hash table with compare-exchange insertion, data-dependent per-lane probe depth, and contended scatter.

The result is a split. On the regular stage, which walks a truncation band and accumulates, all three languages land within a small factor of each other. On the irregular stage, which probes and inserts, Rust stays close to hand-written CUDA C++ while Triton is more than an order of magnitude slower. Language choice is nearly free on the work that is usually benchmarked and expensive on the work that is not.

We attribute both gaps to specific things the languages cannot express, not just ratios. Triton's cost follows from a probe loop that must run to a compile-time bound and from \texttt{tl.atomic\_cas} taking no mask, which forces a scratch structure with no counterpart in CUDA. Rust's cost was invisible in every instruction count: its kernel issues fewer instructions, fewer compare-exchanges and fewer registers at identical occupancy, and was still slower. Hardware counters located it in L1 residency. A GPU-scope atomic load must be coherent across SMs, no NVIDIA L1 is, so the type-correct way to read a shared location bypasses the cache on every access.

Triton's bounded probe is also a correctness problem for fusion: at hash load factors an ordinary depth trajectory reaches, it silently discards blocks and the reconstruction loses patches of surface with nothing reported. We also report a defect found and fixed in \texttt{cuda-oxide} itself, now merged upstream: its scoped atomic load and store could not be called at all in the build mode that produces real kernels.
\end{abstract}

\section{Introduction}

Pick up almost any comparison of GPU programming languages and the benchmark
will be a matrix multiply, a convolution, a stencil, or a reduction. These are
reasonable choices. They are also the workloads every toolchain has been tuned
for, and they share a property that makes them easy: the work per thread is
known before the kernel launches.

This paper measures the other case. A hash-blocked truncated signed distance
function (TSDF) volume, the data structure underneath most real-time 3D
reconstruction systems, is built by a kernel with none of that regularity. Each
point walks a band of voxels, hashes each one to a block coordinate, probes an
open-addressed table for a variable number of steps, and either finds an existing
block or claims one with a compare-exchange. Threads in the same warp do
different amounts of work, contend for the same table slots, and must observe one
another's insertions.

We implement this kernel five times. Two of the implementations are CPU and
third-party baselines used to establish correctness; three are the comparison
proper: CUDA C++, Rust compiled through NVIDIA's \texttt{cuda-oxide}, and
Triton. All three operate on identical device memory with an identical hash
layout, so what varies between them is the source language and nothing else.

\paragraph{Contributions.}
\begin{enumerate}
  \item A measured split between regular and irregular work
        (Section~\ref{sec:results}): the three languages are close on the stage
        that walks and accumulates, and separated by more than an order of
        magnitude on the stage that probes and inserts.
  \item Attribution of both gaps to language-level causes
        (Section~\ref{sec:attribution}), including one that no instruction count
        reveals and that required hardware performance counters to find.
  \item A catalogue of what each language cannot say
        (Section~\ref{sec:expressiveness}), catalogued from what the
        implementation work forced rather than from a wish list.
  \item A measurement methodology and its failure modes
        (Section~\ref{sec:method}), reported because three of the errors it
        caught would have produced a confidently wrong conclusion rather than a
        noisy one.
\end{enumerate}

\paragraph{What this paper is not.} It is not a claim that one language is
better. Rust lands within a few percent of hand-written CUDA C++ on real data
once a single idiom is corrected, and Triton's cost is concentrated in one stage
that a practitioner could write in something else. It is a claim about where the
costs are, and why.

\section{The workload}
\label{sec:workload}

A truncated signed distance function (TSDF) volume stores, at each voxel, the
signed distance to the nearest observed surface, truncated to a band around it.
Fusing a depth image means walking that band along each point's view ray and
accumulating a weighted average. The surface is later recovered as the zero
crossing. It is the representation underneath KinectFusion and most real-time
reconstruction systems since.

A dense volume is unaffordable at useful resolution, so implementations store
only the blocks a surface actually touches, indexed by a spatial hash. That
choice is what makes the kernel irregular, and it is why this workload is the
vehicle for this paper rather than an application of it.

\subsection{Structure}

The volume is divided into blocks of $8^3 = 512$ voxels. A hash table maps block
coordinates to indices in a flat pool of voxel data; five parallel float arrays
hold the truncated distance, the accumulated weight, and three colour channels.
The table is open-addressed with linear probing and sized to the next power of
two above twice the pool capacity, so an occupied table is at most half full.

Each entry is 16 bytes: a 64-bit key and a 32-bit block index, with four bytes
of padding. The key packs the three block coordinates into one word, each axis
biased by $2^{20}$ into 21 bits:
\begin{equation*}
\mathrm{key}(x,y,z) = (x + 2^{20}) \ll 42 \;\vert\; (y + 2^{20}) \ll 21 \;\vert\; (z + 2^{20}).
\end{equation*}
Packing is not a space optimisation. It is what allows a single compare-exchange
to publish an entire coordinate. An earlier layout stored the three axes
separately and published them with a compare-exchange on the first followed by
plain stores to the other two; a reader could then match the first word,
mismatch on the rest, and probe onward, which at GPU thread counts degenerates
into a full-table scan per lookup. The 64-bit key removes that failure by
construction, and every arm inherits it.

\subsection{Two passes, and why}

Integration runs as two kernels over the same points.

\textbf{Allocate} walks the truncation band for each point, computes the block
coordinate of each voxel it touches, and ensures a block exists for it. This is
the irregular stage: threads probe for a data-dependent number of steps, race
for the same slots, and must observe one another's insertions.

\textbf{Update} walks the same band and accumulates into blocks that now exist.
This is the regular stage: a lookup that almost always hits on the first probe,
followed by five atomic additions per voxel.

Accumulation stores weighted \emph{sums} rather than running means. A running
mean requires reading the current value, combining, and writing back, which no
single atomic can make safe across five arrays; sums commute, so plain atomic
additions suffice and the arms need not agree on a locking scheme. The mean is
recovered at extraction.

Both passes apply the same occlusion test. Applying it in only one leaves blocks
that are allocated but never receive a contribution, which does not change the
extracted surface but does change the block count, and the block count is one of
the cross-arm correctness checks.

\subsection{The insert protocol}

Listing~\ref{lst:insert} is the CUDA C++ insert, and is the shape every arm
implements.

\begin{lstlisting}[caption={The insert protocol. Both the probe loop and the
publication window shape the comparison in Section~\ref{sec:attribution}.},
label={lst:insert}]
for (probe = 0; probe < size; ++probe) {
    entry = table[(slot + probe) & mask];
    key   = entry.key;                      // plain, cacheable load
    if (key == want) {                      // block already exists
        idx = entry.block_idx;
        while (idx < 0)                     // wait for the winner to publish
            idx = *(volatile int*)&entry.block_idx;
        return idx;
    }
    if (key == EMPTY) {
        prev = atomicCAS(&entry.key, EMPTY, want);
        if (prev == EMPTY) {                // we won the slot
            idx = atomicAdd(block_count, 1);
            block_coord[idx] = {bx, by, bz};
            __threadfence();                // coords visible before the index
            atomicExch(&entry.block_idx, idx);
            return idx;
        }
        // lost the race; fall through and re-examine this slot
    }
}
\end{lstlisting}

Two features of this protocol drive everything that follows.

\paragraph{The probe is unbounded.} It runs until it finds the key or an empty
slot, so its trip count is data-dependent and varies per lane. Section
\ref{sec:expressiveness} shows this is not expressible in Triton at any price.

\paragraph{Publication is two steps.} The key is published by a compare-exchange
and the index by a later store, so a reader can observe a key whose index is
still absent and must wait. The window exists because a block's \emph{identity}
is known before insertion while its \emph{location} is assigned by a counter and
known only after winning. Publishing both atomically would need the location
first; the location is only yours if the exchange succeeds. That circularity, and
two attempts to remove it, are discussed in Section~\ref{sec:attribution}.

\subsection{What is excluded, and why}

Surface extraction is not part of the comparison. It exists once, in CUDA C++,
and all three GPU arms call it, so measuring it three times compares nothing.
Making it a real comparison would mean writing marching tetrahedra twice more,
and that work would not extend the argument: it is a regular kernel, and the
comparison already contains one in the update stage.

Investigating the decision also corrected a number. The extraction stage had been
reported at 1.825\,ms and as several times the cost of the whole integrate path.
Splitting the call showed that 82\% of that measurement on one GPU and 97\% on the
other was a device-to-host transfer over PCIe links of differing width, not
compute. The kernel and its allocation are 0.319\,ms, comparable to integrate
rather than several times it, and the striking cross-device difference was the
motherboard.

\begin{figure}[h]
\centering
\begin{minipage}{0.235\linewidth}\centering
\includegraphics[width=\linewidth]{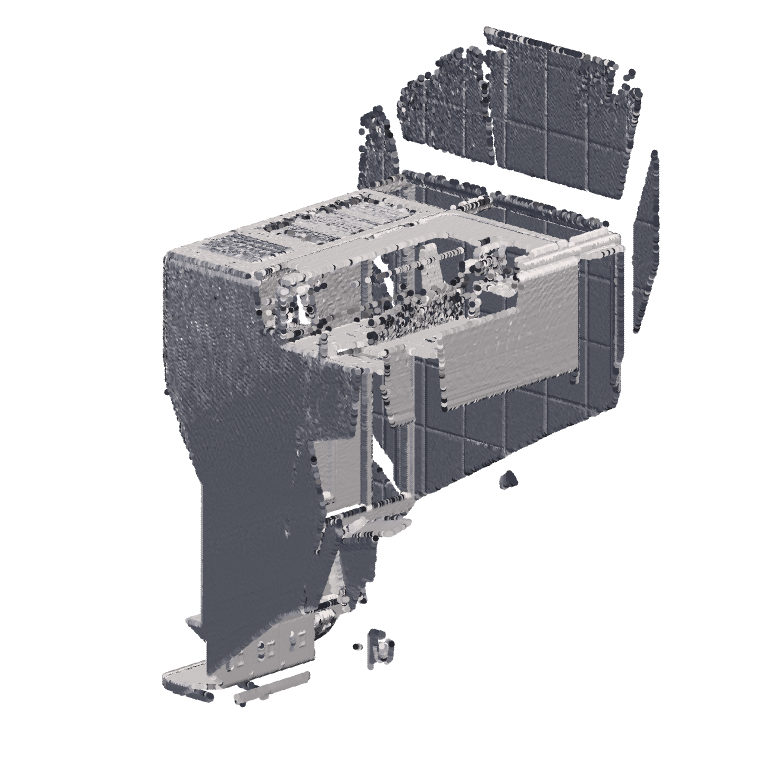}\\{\scriptsize RetroOffice, warm}
\end{minipage}\hfill
\begin{minipage}{0.235\linewidth}\centering
\includegraphics[width=\linewidth]{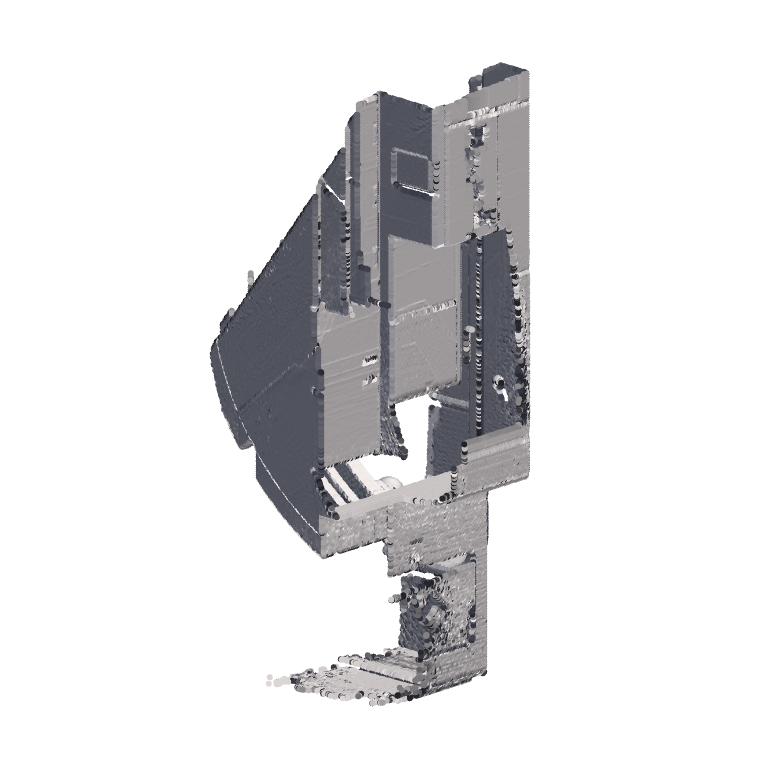}\\{\scriptsize AmericanDiner, warm}
\end{minipage}\hfill
\begin{minipage}{0.235\linewidth}\centering
\includegraphics[width=\linewidth]{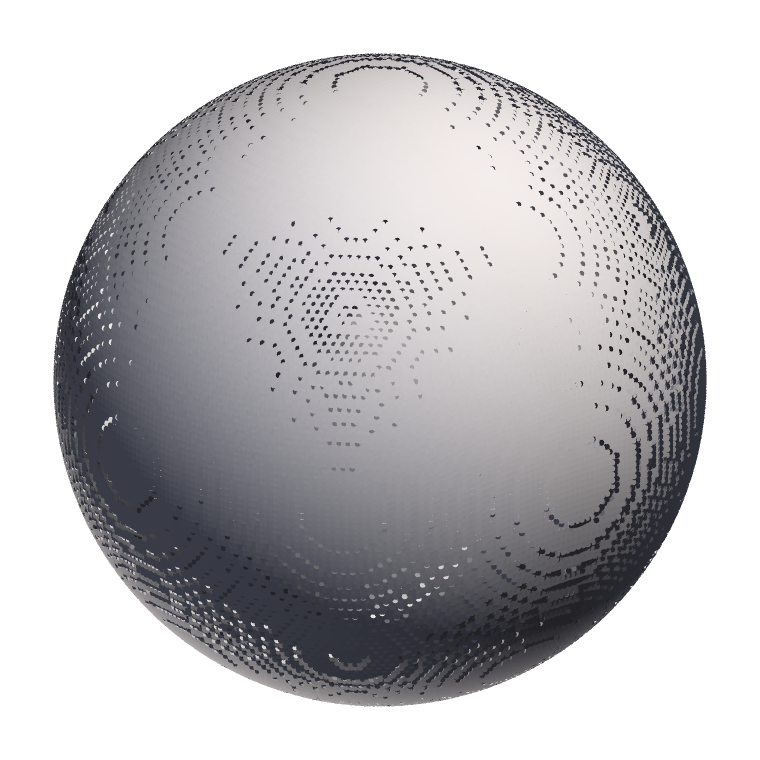}\\{\scriptsize sphere, 320k points}
\end{minipage}\hfill
\begin{minipage}{0.235\linewidth}\centering
\includegraphics[width=\linewidth]{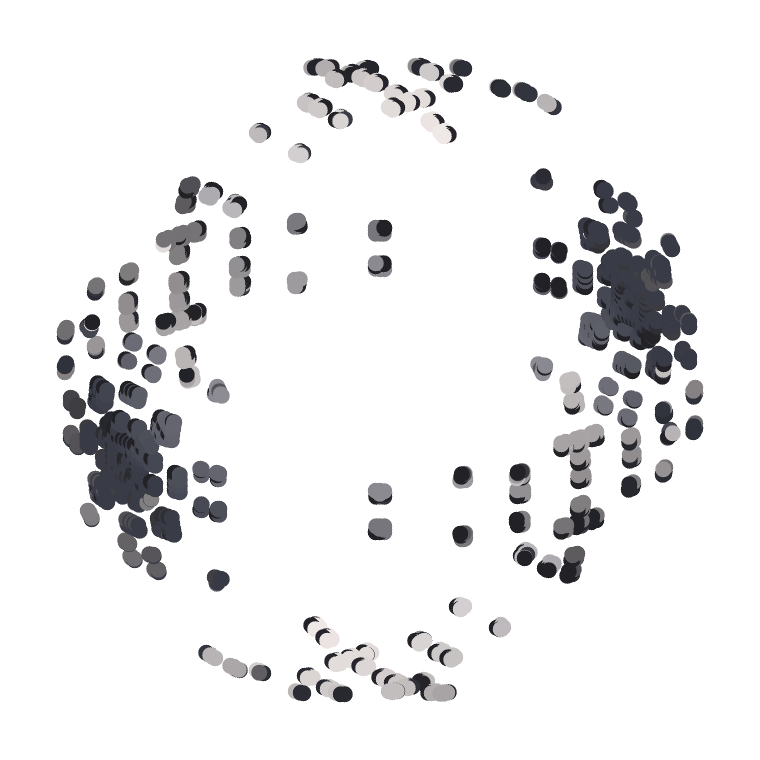}\\{\scriptsize sphere, 20k points}
\end{minipage}

\vspace{6pt}
\begin{minipage}{0.235\linewidth}\centering
\includegraphics[width=\linewidth]{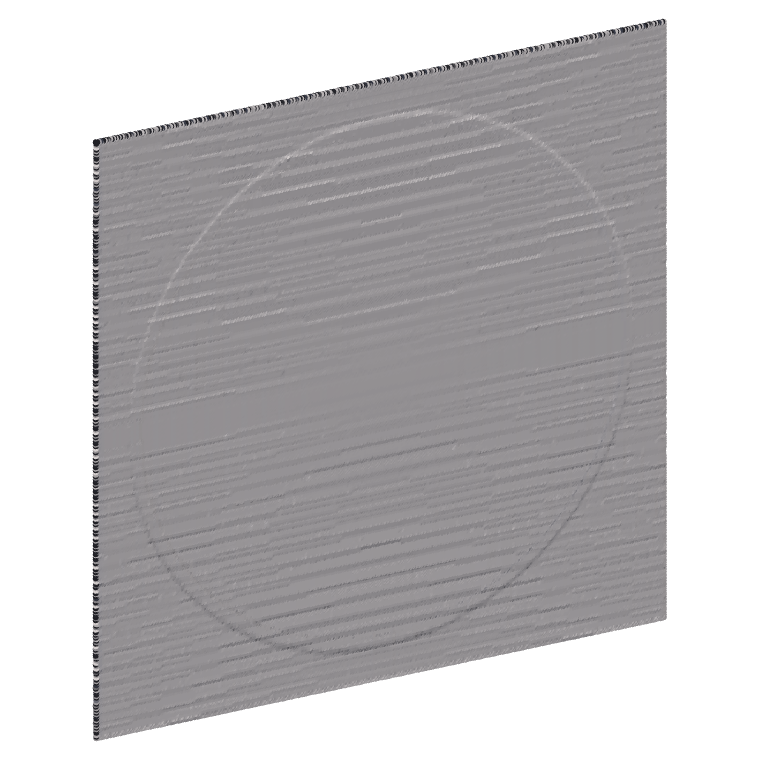}\\{\scriptsize plane, 320k points}
\end{minipage}\hfill
\begin{minipage}{0.235\linewidth}\centering
\includegraphics[width=\linewidth]{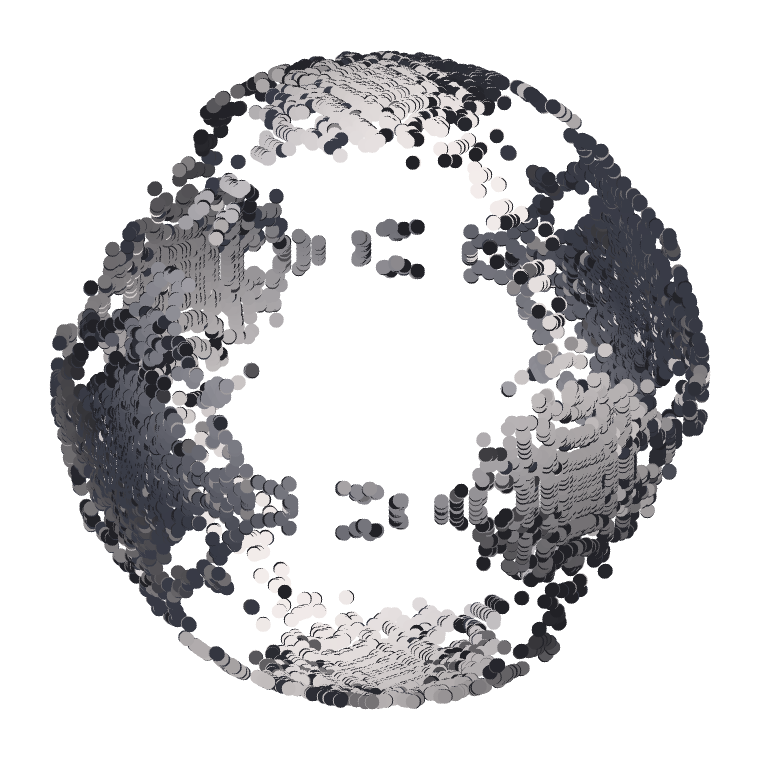}\\{\scriptsize sphere, 2\,m radius}
\end{minipage}\hfill
\begin{minipage}{0.235\linewidth}\centering
\includegraphics[width=\linewidth]{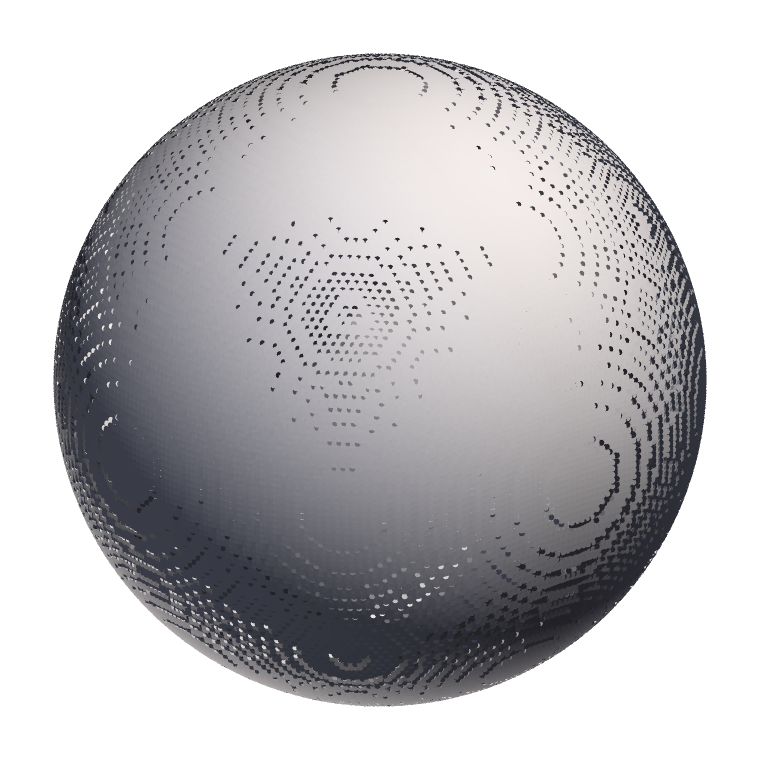}\\{\scriptsize load factor 0.283}
\end{minipage}\hfill
\begin{minipage}{0.235\linewidth}\centering
\includegraphics[width=\linewidth]{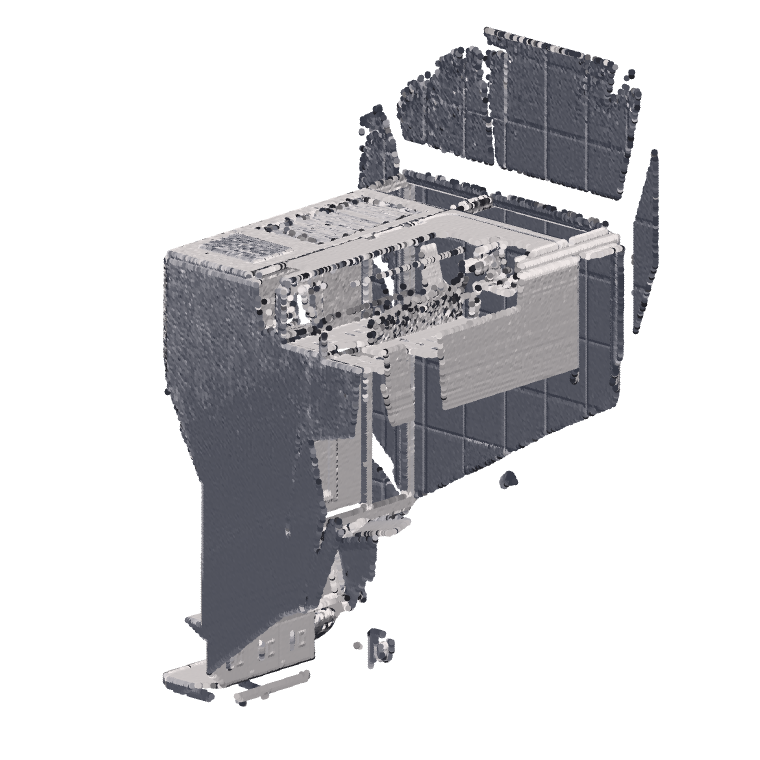}\\{\scriptsize real, load factor 0.437}
\end{minipage}
\caption{What the cells actually build, rendered from the fused volume: one lit
point per occupied voxel, shaded by the surface normal the TSDF supplies. These
are the reconstructions the timings come from, not illustrations of them. The
two rooms are different environments, which is what the real-data axis varies
beyond camera path. The 20k-point cell is the sweep's outlier in every table,
and the reason is visible here: at that density the volume barely reaches the
weight threshold anywhere, so there is almost no work for the kernel to do and
a fixed per-launch cost dominates whatever the language contributes.}
\label{fig:scenes}
\end{figure}

\subsection{Why this workload}

The properties that make TSDF fusion awkward are exactly the ones absent from
the benchmarks these languages are usually compared on:

\begin{itemize}
  \item work per thread is data-dependent and not known at launch;
  \item threads in a warp diverge on both probe depth and outcome;
  \item correctness depends on compare-exchange and on one thread observing
        another's publication;
  \item the access pattern is a scatter through a hash, with no tile structure
        to exploit.
\end{itemize}

None of this is exotic. It is what any GPU hash table, sparse structure builder,
free-list allocator, or dynamic work queue looks like. The results in
Section~\ref{sec:results} should be read as applying to that class, not to
reconstruction specifically.

\section{Implementations}
\label{sec:impl}

Five implementations exist. Three are the comparison; two are there to make the
three trustworthy.

\begin{description}
  \item[A1, Open3D.] A third-party voxel block grid, used as an external
        reference for the algorithm's shape. Not timed against the others.
  \item[A2, CPU C++.] The same algorithm on the host, in C++20. Establishes that
        the reference results are not a property of any GPU implementation.
  \item[A3, CUDA C++.] The performance reference.
  \item[A4, Rust.] Compiled to PTX through \texttt{cuda-oxide}.
  \item[A5, Triton.] Two variants: one where allocation runs on A3's kernel and
        only the update stage is Triton, and one where both stages are Triton.
        The difference between them isolates the cost of the irregular stage
        without relying on cross-arm comparison.
\end{description}

\subsection{What is held fixed}

The three compared arms share the device memory layout exactly: the same 16-byte
hash entry, the same packed key, the same multiply-xor hash, the same pool
indexing, the same 256-thread launch geometry, and the same truncation band. Each
arm is handed pointers into memory allocated once by the harness, through a C
ABI, and none of them owns or reshapes it.

This is deliberate and it is a constraint on the result. Holding the layout fixed
means the comparison measures how each language expresses one algorithm, not
which algorithm each language makes convenient. A Triton implementation free to
choose its own data structure would look different and probably better; that is a
different paper, and a less controlled one.

The hash function in particular must be bit-identical across arms. A different
mix changes probe sequences, which changes which blocks share cache lines, which
would appear in the timings as a language difference.

\subsection{What is not shared}

No code is reused between arms. Each is written from the algorithm rather than
ported from another arm's source.

The original design had a vendored reference implementation that the other arms
were ported from and then checked against. We removed it. Certifying an arm
against the implementation it was derived from is weakly circular: a shared
misunderstanding of the algorithm passes the check. The correctness design in
Section~\ref{sec:method} replaces it with closed-form geometry, which has no such
failure mode.

\subsection{Correctness}

Every arm must reproduce a sphere of known radius to within a quarter of a voxel,
agree with the reference on the number of allocated blocks exactly, and drop no
points. In the final state the three GPU arms agree with each other to
$0.000000000$\,m mean surface distance and $5.5 \times 10^{-8}$\,m Hausdorff
distance, on 846{,}288 extracted vertices.

Timing is gated on those checks, and the sweep additionally discards any cell in
which an arm drops a block or disagrees on block count. That gate is not
ceremonial: it is what detected the silent data loss described in
Section~\ref{sec:expressiveness}, in which one arm quietly built 13 fewer blocks
than the others at a load factor no previous benchmark had reached.

\subsection{Completeness}

No arm is a sketch. Each implements the full algorithm: the truncation-band walk,
the occlusion test, hash insertion with the publication protocol of
Listing~\ref{lst:insert}, pool-exhaustion reporting, and weighted accumulation
across all five voxel arrays. Each passes the same correctness gate. The
differences reported in Section~\ref{sec:results} are therefore differences in
how the same work is expressed, not in how much of it was done.

We deliberately do not compare line counts. The Rust source carries the
measurement variants described in Section~\ref{sec:attribution} alongside the
production kernels, the Triton source carries two arm variants, and the CUDA
source carries host-side management the others receive through the C ABI. Any
line-count comparison across those would measure our file organisation rather
than the languages.

\section{Methodology}
\label{sec:method}

Most engineering benchmark papers are rejected for sloppy measurement rather
than for uninteresting numbers. We describe the harness in detail, and then
describe the errors it did \emph{not} prevent, because the second list is more
useful than the first.

\subsection{Correctness gates timing}

No arm is timed until it is known to compute the right answer, and the checks
are ordered by how much they can catch.

\textbf{Analytic first.} Points are sampled on a sphere of known radius, so the
extracted surface has a closed-form expected position. This is the only tier
that can catch an error all arms share.

\textbf{Third party second.} Open3D's voxel block grid provides an external
reference for the algorithm's shape.

\textbf{Cross-arm last.} The arms are compared to one another, at 0.000000000\,m
mean surface distance in the final state.

The order is not decoration. A half-voxel bias, in which the signed distance was
evaluated at the ray sample that selected a voxel rather than at the voxel
centre, passed the plane test and every cross-arm comparison, because all arms
had inherited it. It was caught only by the sphere, where it appeared as a
uniform 0.005\,m radius error; correcting it improved agreement to 0.000015\,m, a
factor of 340. Had the project been built on cross-arm agreement alone, every
number in it would have been wrong in the same direction and the agreement would
have looked like evidence.

\subsection{What the harness defends against}

Each of the following exists because its absence produced a wrong number at some
point during the work.

\textbf{Amortised launches.} Each timed window contains several kernel launches
rather than one, divided out afterwards. Timing a single launch measures the
host's wait for completion as much as the kernel: the driver spins briefly and
then blocks, and the scheduler wakeup of roughly 130\,\textmu s lands unevenly
across arms depending on how much host work each does after enqueuing. This
produced a spurious $4\times$ tail on one arm that a separate probe showed was
entirely in the wait, with the kernel itself flat.

Allocation additionally needs distinct volumes per launch: after the first
launch the table is populated, so re-launching into the same volume would measure
a lookup-only fast path.

\textbf{Interleaved and rotated arm order.} Arms take turns within each
repetition and the starting arm rotates. Running all of one arm and then all of
another makes clock ramp indistinguishable from a language difference; a single
session was observed going from 1575\,MHz to 2842\,MHz, which is larger than
several of the effects being measured. A per-arm first-half against second-half
comparison is printed as a check that the interleaving worked.

\textbf{Exclusive device.} The harness refuses to run when another process holds
significant memory on the target GPU. An unrelated job sharing the device once
made one arm measure 4.455\,ms against its true 0.044\,ms, a $100\times$ error
that looked like a plausible result.

\textbf{Distributions, not means.} Median, 95th and 99th percentiles per stage,
with the repeat count stated. Clocks and temperatures are recorded per cell
rather than once per run, because a later cell running hotter than an earlier one
is an artefact that is only visible if sampled alongside the timings.

\subsection{The sweep}

Every number comes from a matrix rather than a single scene, because a ratio
measured once is a property of that scene's contention pattern as much as of the
language. Four axes, chosen so that each answers a stated question rather than
filling a table: point count at fixed geometry, which raises contention while
holding hash work constant; extent at fixed angular resolution, which is the
opposite lever; hash load factor; and scene shape, including two cells of real
depth data.

The matrix runs on two GPUs and three times over, and cells are discarded if any
arm drops a block or disagrees with the reference on block count. Automatic
reductions in batch size, when a cell's volumes will not fit in memory, are
reported rather than silently applied, because a cell measured at a different
batch size is not comparable to one that was not.

\subsection{What went wrong anyway}

Roughly a dozen errors were caught during this work. Most produced a wrong
magnitude. Three produced a reversed or null conclusion, and two of those survived
long enough to be written up.

\paragraph{A workload asymmetry made the slower arm look faster.} Colour
accumulation was implemented in three arms and not the fourth, a 40\% difference
in atomic traffic. Rust appeared $2\times$ faster than CUDA C++.

\paragraph{A device axis was silently fake.} One arm's loader hardcodes device
zero, and creating a driver-API context makes it current for the whole process,
so an explicit device selection was ignored by every arm. The first two-device
sweep reported a clean null result, identical performance across a $2.33\times$
difference in streaming multiprocessor count, while the second card sat idle at
8\,W. That is a striking and publishable-looking finding with a ready
explanation, and nothing in the timings looked anomalous. It was caught only
because the null was surprising enough to prompt a check of the card's power
draw. The harness now selects the device by masking the process to one GPU and
verifies the resulting context, because the runtime's notion of the current
device and the driver's can differ.

\paragraph{A single pass was not reproducible.} One cell reported 0.047\,ms in
one pass and 0.060\,ms in each of three reruns, a 30\% swing. No within-pass
statistic separates that case: flagging on the ratio of the 95th percentile to
the median fires on eight of ten cells and misses this one. Only repetition of
the whole matrix detects it, which is why the sweep runs three passes and reports
medians.

A fourth is worth listing because it was a stage-level rather than an arm-level
error, and because it shows the same failure at a different granularity: the
extraction stage's reported time turned out to be mostly bus traffic rather than
compute (Section~\ref{sec:workload}).

\paragraph{Two more silent-measurement defects, found by widening the sweep.}
Adding real-data cells surfaced two failures of the same kind as the fake
device axis, and we record them for the same reason. First, a cell that sizes
its pool from the measured block count sized it from the frame alone, ignoring
the warm pre-fill, so the pool was exhausted before the timed window opened and
the cell measured pool exhaustion rather than table contention; because an
unbounded probe against a full table scans every slot for every point, a cell
that should run in two seconds ran for over half an hour, which is the only
reason we noticed. Second, the Triton cubin path is resolved relative to the
working directory, and the sweep driver runs from the build directory, so an
override that is correct from the repository root silently drops the Triton
arm: a complete sweep ran, reported ``0 invalid'', and compared two arms where
three were asked for. The harness now refuses to run with an arm missing unless
told explicitly to allow it. A third instance, the same day, was a sweep run at
the historical scratch-region size, which costs the Triton arm a factor of
three to four; that value is no longer the default, because a default is what
you measure by forgetting a flag.

\subsection{What this suggests}

The common thread is that all three reversing errors were failures of
\emph{comparability} rather than of precision. The two sides were not doing the
same work, or were not running on the same machine, or were not being summarised
by a statistic that survived repetition. None of them would have been caught by
running more repetitions or reporting tighter error bars.

A language comparison's credibility rests entirely on those three properties,
and in this project each of them turned out to be something the harness had to
verify rather than assume. We suggest that any such comparison state explicitly
how it established them.

\section{Results}
\label{sec:results}

Table~\ref{tab:stage} is the paper in one table.

\begin{table}[h]
\centering
\begin{tabular}{llrr}
\toprule
Stage & Character & Rust / CUDA C++ & Triton / CUDA C++ \\
\midrule
allocate & irregular: probe, CAS, publication & 1.02--3.34 & 11.2--31.6 \\
update & regular: walk and accumulate & 0.96--1.19 & 1.1--2.6 \\
\bottomrule
\end{tabular}

\caption{Ratio to hand-written CUDA C++ by stage, across every recorded cell of
the sweep and both GPUs. The regular stage separates the languages by a small
factor; the irregular stage separates them by more than an order of magnitude.
The wide ends of both ranges are the smallest cell, where the kernel is short
enough that launch effects dominate; see Section~\ref{sec:results}.}
\label{tab:stage}
\end{table}

\begin{table}[h]
\centering
\small
\begin{tabular}{llrrrrr}
\toprule
Workload & GPU & CUDA C++ & \multicolumn{2}{c}{Rust} & \multicolumn{2}{c}{Triton} \\
\cmidrule(lr){4-5}\cmidrule(lr){6-7}
 & & ms & ms & $\times$ & ms & $\times$ \\
\midrule
RetroOffice P000, warm & RTX 5070 Ti & 0.388 & 0.396 & 1.02 & 1.132 & 2.92 \\
RetroOffice P000, warm & RTX 5060 & 0.763 & 0.768 & 1.01 & 2.229 & 2.92 \\
RetroOffice P000, cold & RTX 5070 Ti & 0.472 & 0.486 & 1.03 & 1.215 & 2.57 \\
RetroOffice P000, cold & RTX 5060 & 0.905 & 0.915 & 1.01 & 2.335 & 2.58 \\
AmericanDiner P000, warm & RTX 5070 Ti & 0.400 & 0.417 & 1.04 & 1.143 & 2.86 \\
AmericanDiner P000, warm & RTX 5060 & 0.780 & 0.796 & 1.02 & 2.105 & 2.70 \\
Plane, 320k points & RTX 5070 Ti & 0.363 & 0.379 & 1.04 & 0.875 & 2.41 \\
Plane, 320k points & RTX 5060 & 0.673 & 0.686 & 1.02 & 1.665 & 2.47 \\
Sphere, 1.28M points & RTX 5070 Ti & 0.975 & 1.023 & 1.05 & 2.982 & 3.06 \\
Sphere, 1.28M points & RTX 5060 & 1.709 & 1.737 & 1.02 & 5.707 & 3.34 \\
Sphere, 320k points & RTX 5070 Ti & 0.278 & 0.308 & 1.11 & 0.828 & 2.98 \\
Sphere, 320k points & RTX 5060 & 0.511 & 0.538 & 1.05 & 1.526 & 2.99 \\
\bottomrule
\end{tabular}

\caption{Full integrate path, allocate plus update, median of three passes.
Real depth data (TartanAir) and synthetic scenes, on two GPUs differing
$2.33\times$ in streaming multiprocessor count.}
\label{tab:totals}
\end{table}

\subsection{The split}

\begin{figure}[h]
\centering
\includegraphics[width=\linewidth]{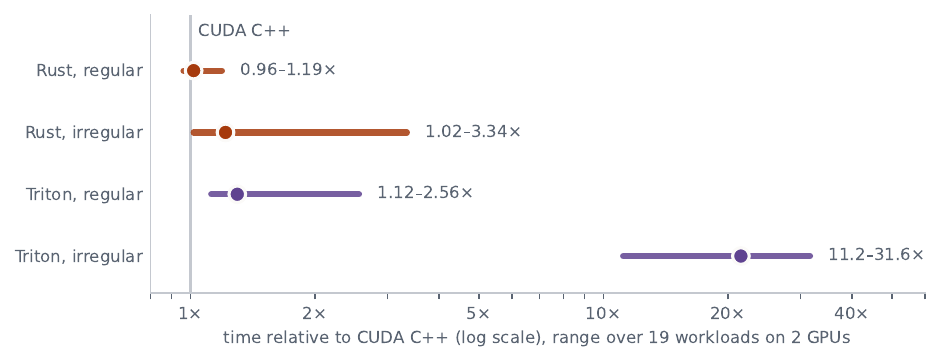}
\caption{The result. The same two implementations, priced separately on the
regular stage and the irregular one, as a range over all nineteen workload cells
with the median marked. Both arms sit close to CUDA C++ on the stage that walks
a band and accumulates; on the stage that probes and inserts, Triton is more
than an order of magnitude behind and Rust's cost becomes strongly
workload-dependent. Log scale.}
\label{fig:stages}
\end{figure}

Figure~\ref{fig:stages} and Table~\ref{tab:stage} are the result. On the update stage, which walks a band and
accumulates, the three languages differ by a small factor. On the allocate
stage, which probes and inserts, Rust remains close to CUDA C++ while Triton is
more than an order of magnitude behind.

The real-depth cells sit at the \emph{worse} end of Triton's range rather than
the middle of it: across the nine valid real cells its allocate ratio spans
$17.3$--$31.6\times$ with a median of $25.2\times$, against $11.2$--$31.6\times$
and a median of $21.5\times$ over the sweep as a whole. Rust does not show the
same effect, spanning $1.12$--$1.41\times$ on real data against $1.02$--$3.34$
overall. Synthetic scenes therefore understate the cost of the construct
Triton lacks, which matters because a synthetic scene is what a language
comparison would ordinarily be run on.

The wide ends of both ranges belong to the smallest cell, 20{,}000 points, where
the kernel runs in a few microseconds and a fixed per-launch cost dominates
anything the language does. We report the full range rather than excluding the
cell, and Table~\ref{tab:cells} shows where it sits.

On the full integrate path (Table~\ref{tab:totals}) the allocate stage is roughly
a tenth of the work, so the totals are dominated by the regular stage: Rust lands
within 1--3\% of CUDA C++ on real depth data and Triton at $2.6$--$2.9\times$. A
practitioner reading only the totals would conclude that Rust is free and Triton
is expensive; reading the stages shows that both conclusions come from one
kernel.

\subsection{The sweep, and what each axis shows}

\begin{figure}[h]
\centering
\includegraphics[width=\linewidth]{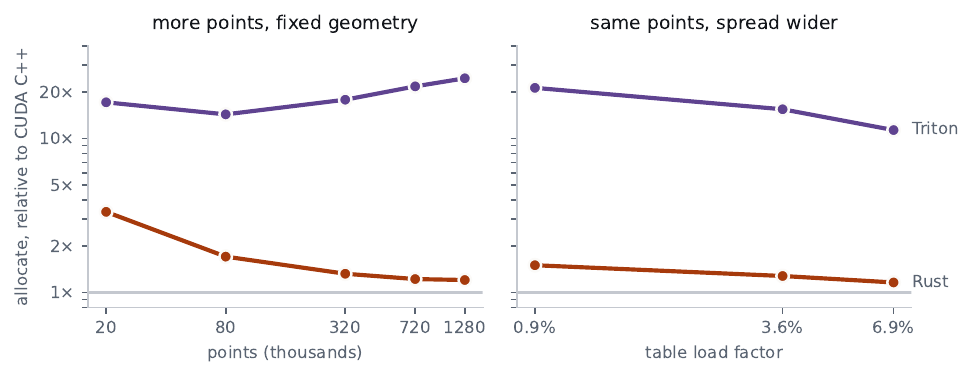}
\caption{Two levers that move contention in opposite directions, on the
RTX~5070~Ti. Adding points over fixed geometry raises the number of threads
contending for each table slot, and Triton's allocate cost rises with it;
spreading the same points over more blocks lowers contention, and Triton's cost
falls. A cost driven by problem size would slope the same way in both panels.
Rust falls in both, which is the shape of a fixed per-launch cost being
amortised rather than a contention cost. Log--log.}
\label{fig:axes}
\end{figure}

\begin{table}[h]
\centering
\small
\begin{tabular}{llrrrr}
\toprule
Cell & Axis & Points & Load & Rust & Triton \\
\midrule
base & baseline & 320,000 & 0.035 & 1.32 & 17.9 \\
\addlinespace
pts-20k & points & 20,000 & 0.034 & 3.34 & 17.2 \\
pts-80k & points & 80,000 & 0.035 & 1.71 & 14.4 \\
pts-720k & points & 720,000 & 0.035 & 1.22 & 21.8 \\
pts-1280k & points & 1,280,000 & 0.035 & 1.21 & 24.6 \\
\addlinespace
r-0.25 & extent & 320,000 & 0.009 & 1.50 & 21.4 \\
r-1.0 & extent & 320,000 & 0.036 & 1.28 & 15.5 \\
r-2.0 & extent & 320,000 & 0.069 & 1.16 & 11.4 \\
\addlinespace
plane-320k & shape & 320,356 & 0.032 & 1.21 & 17.1 \\
\addlinespace
lf-sparse & loadfactor & 320,000 & 0.018 & 1.49 & 17.7 \\
\addlinespace
tartan-p001 & real & 374,692 & 0.043 & 1.33 & 31.6 \\
diner-p000 & real & 393,624 & 0.094 & 1.33 & 26.4 \\
tartan-p004 & real & 400,248 & 0.066 & 1.30 & 29.6 \\
tartan-p005 & real & 401,678 & 0.108 & 1.30 & 24.8 \\
tartan-p003 & real & 405,269 & 0.114 & 1.24 & 28.4 \\
tartan-cold & real & 407,444 & 0.078 & 1.23 & 17.8 \\
tartan-warm & real & 407,444 & 0.109 & 1.24 & 28.5 \\
diner-p002 & real & 409,600 & 0.021 & 1.41 & 30.5 \\
diner-p003 & real & 409,600 & 0.117 & 1.33 & 28.6 \\
\bottomrule
\end{tabular}

\caption{Allocate ratio to CUDA C++ per cell, on the RTX 5070 Ti. Ratios, not
times, because the axes change the absolute work by two orders of magnitude. The
trends within each axis group are the evidence discussed below.}
\label{tab:cells}
\end{table}

Table~\ref{tab:cells} groups the cells by axis. Two of the axes were chosen as
opposite levers, and they move the Triton ratio in opposite directions, which is
the strongest evidence in the paper that the mechanism in
Section~\ref{sec:attribution} is the right one.

\paragraph{Point count, at fixed geometry.} Raising the point count over the same
surface raises contention while leaving the number of blocks unchanged. Triton's
ratio rises with it, from $14.2\times$ to $24.7\times$, because every additional
lane that resolves early still issues a compare-exchange it cannot mask off.
Rust's ratio \emph{falls}, from $3.34\times$ to $1.21\times$, which is the
signature of a fixed cost being amortised rather than a multiplier: the absolute
difference is roughly constant while the work grows $64\times$.

\paragraph{Extent, at fixed angular resolution.} Spreading the same points over a
larger sphere raises the number of blocks while lowering per-voxel contention.
Triton's ratio falls, from $21.3\times$ to $11.4\times$: with more blocks, fewer
lanes resolve on the first probe, so the wasted work is a smaller share of the
total. Two levers, opposite directions, one mechanism.

\paragraph{Load factor.} The sharpest prediction of the Triton explanation is
that its ratio should shrink as the table fills, because CUDA C++ pays actual
probe depth while Triton pays a fixed bound. We could not test it. The table is
sized at twice the pool and the pool must hold every block, so the reachable load
factor caps at 0.283, over which range CUDA's expected probe count only moves
from 1.02 to 1.39. The lever is too weak, and this is a limitation of the data
structure rather than of the sweep (Section~\ref{sec:threats}).

\paragraph{Real data.} Two cells unproject ground-truth depth from a TartanAir
sequence. They validate the synthetic sweep rather than contradicting it: the
ratios land inside the synthetic range and within a fraction of a point of the
baseline cell. The warm cell, which integrates a frame into a volume already
holding the preceding eight, is the regime a live pipeline is in for every frame
after the first, and it is where Triton is worst in the entire sweep at
$28.6\times$. In that regime most probes hit an existing key and exit
immediately; CUDA C++ and Rust take that exit and Triton, which cannot, does not.

\subsection{The device axis}

The matrix runs on two GPUs differing $2.33\times$ in streaming multiprocessor
count at identical code generation, since both are the same architecture. All
three arms now scale with the machine, between $1.1\times$ and $2.2\times$
against an available $2.33\times$, with the low ends again the smallest cell.

This axis was not always uninteresting. Before a fix described in
Section~\ref{sec:attribution}, Triton's allocate did not scale at all, at
$0.89\times$ to $1.07\times$: adding forty streaming multiprocessors bought it
nothing, and on some cells the narrower card was faster. That failure is
invisible on a single device, which reports a plausible time and no indication
that the kernel has stopped responding to hardware.

The axis also serves as a check on the ratios themselves. Per cell, the two cards
agree closely: Triton at $17.9\times$ on both at the baseline, Rust at
$1.01$--$1.03\times$ on both with real data. A ratio that survives a change of
machine balance is a better candidate for a language property than one that does
not, and an earlier version of these numbers halved between the two cards, which
was a symptom rather than a finding.

\section{Attribution}
\label{sec:attribution}

Reporting a ratio without a mechanism is what makes most language comparisons
hard to cite. This section attributes both gaps, and reports the part of one that
remains unattributed.

\subsection{Triton: two costs that compound}

\paragraph{The probe loop must run to a fixed bound.} Triton has no per-lane
early exit. A lane that resolves on the first probe cannot leave the loop; it can
only be masked out of subsequent work while the loop continues. The bound is a
\texttt{tl.static\_range} trip count and therefore a compile-time constant, so
every lane pays worst-case probe depth. Measured by varying the bound, the cost
is linear in it: $8.38\times$ for an $8\times$ increase, confirmed again at a
bound of 32.

\paragraph{\texttt{tl.atomic\_cas} takes no mask.} Unlike
\texttt{tl.atomic\_add}, the compare-exchange has no mask parameter, so lanes
that have already resolved cannot be suppressed and still issue an exchange
somewhere. They must be aimed at a scratch address holding a sentinel that can
never satisfy the compare. The programmer thereby acquires a structure the CUDA
version does not have, along with its sizing and its indexing.

The two compound: the bound sets how many iterations run, and the mask
restriction sets what each wasted iteration costs. Replacing the exchange with a
plain load, as a control, leaves 0.090\,ms against 2.04\,ms, a factor of 113, so
essentially the entire gap is on this path rather than in code generation or
memory layout.

\subsection{Triton: a failure invisible on one device}

The scratch region was originally indexed by lane, giving 256 addresses. Since
lane indices repeat across the grid, \emph{every} program used the same 256
addresses: the number of contenders grows with the grid while the address set
does not.

The consequence was not primarily a slowdown. It was that the kernel stopped
responding to hardware. Across a $2.33\times$ difference in streaming
multiprocessor count, its allocate stage ran at $0.89$--$1.07\times$, and on some
cells the \emph{narrower} card was faster, which is the signature of a
serialising address set: fewer contenders is marginally better. The other two
arms tracked the machine at $1.8$--$2.1\times$ over the same cells.

Indexing the region per program instead restores scaling to
$1.76$--$2.07\times$ and makes the kernel $3.5\times$ faster, from 1.812\,ms to
0.524\,ms. Saturation occurs before every lane has a private address, so
contention is relieved statistically rather than eliminated.

This correction cost the headline. An earlier version of this work reported
Triton's allocate gap as $73\times$; sized correctly it is $17.9\times$. Most of
the difference was a badly-chosen parameter of the workaround, not the language.
We report the smaller number as the stronger claim, because it is
device-independent where the larger one halved between cards.

The general lesson is the one a single-GPU benchmark cannot reach: the missing
mask forces a structure whose bad settings are invisible in a throughput number.
One GPU reports 1.8\,ms and looks reasonable.

\subsection{Rust: eleven hypotheses and a profiler}

\begin{table}[h]
\centering
\small
\begin{tabular}{lrrr}
\toprule
Counter & CUDA C++ & Rust & Rust, fixed \\
\midrule
static SASS instructions & 520 & 528 & 528 \\
branches (BRA/BSSY/BSYNC/BRX) & 72 & 63 & 63 \\
global memory ops (LDG/STG) & 11 & 6 & 6 \\
registers per thread & 34 & 28 & 28 \\
register spills & 0 & 0 & 0 \\
dynamic compare-exchange attempts & 28,558 & 25,026 & 25,026 \\
blocks built & 1,160 & 1,160 & 1,160 \\
achieved occupancy & 84.30 & 84.27 & 84.27 \\
active threads per warp & 26.95 & 28.18 & 28.18 \\
\addlinespace
warp cycles per issued instruction & 19.73 & 23.21 & -- \\
L1 sector hit rate & 55.86 & 28.92 & 54.12 \\
L2 sectors & 568,655 & 964,189 & 552,775 \\
long\_scoreboard stall & 12.02 & 15.63 & 11.06 \\
duration & 44.96 & 52.16 & 42.18 \\
\bottomrule
\end{tabular}

\caption{Allocate kernel counters, 320k points, RTX 5070 Ti. Above the rule are
counts of work: Rust does less on every one except static instruction count,
where it is within 2\%, and it was still slower. Below the rule are the counters
that show why. ``Rust, fixed'' is after the change described in this section; the
dash marks a figure not recorded for that build.}
\label{tab:counters}
\end{table}

Rust's allocate gap resisted attribution longer than any other result here, and
the reason is visible in Table~\ref{tab:counters}: it is not a
\emph{quantity-of-work} problem. The Rust kernel issues fewer branches, fewer
global memory operations, and fewer dynamic compare-exchanges, from a smaller
register budget, at identical occupancy and with less warp divergence, producing
an identical result. It does less work, more slowly.

Eleven hypotheses were tested and eliminated: register pressure, \texttt{clamp}
semantics, volatile qualification, static instruction count, dynamic
compare-exchange count, memory-ordering scope, the shared block counter, the
publication fence, the publication store, the publication wait, and host-side
launch overhead. Several were eliminated by experiments that required
substantial work, including one that needed a patch to the Rust compiler backend
before it could be run at all (Section~\ref{sec:expressiveness}).

Hardware counters located it in one run. Across every stall reason the profiler
reports, \texttt{long\_scoreboard}, the wait for a long-latency memory operation
to return, was the \emph{only} one that worsened, and it worsened by more than the
whole gap; every other reason favoured Rust. The cause was L1 residency:
identical requests over identical sectors, but $1.70\times$ the sectors pushed
through to L2, and an L1 hit rate of 28.9\% against 55.9\%.

\paragraph{The mechanism.} Two loads. The Rust implementation read the probe key
and the published block index with a scoped atomic load; the CUDA implementation
reads both plainly.

A note on sequence, because Section~\ref{sec:expressiveness} reports that the
scoped atomic load could not be called at all in the version of
\texttt{cuda-oxide} we started from. It could not, and the Rust arm originally
used a volatile read instead. That construct \emph{also} bypasses L1, and more
expensively, since Rust's volatile semantics lower to a system-scope access. We
patched the compiler in order to test a different hypothesis, which failed; the
patch then made the scoped atomic load usable, and it is the version reported
here. The relevant point survives the detour and is arguably strengthened by it:
\emph{both} constructs Rust offers for reading a location another thread writes
bypass the cache, and the fix was to use neither. \textbf{A GPU-scope atomic load must be coherent across
streaming multiprocessors, and no NVIDIA L1 cache is coherent across streaming
multiprocessors, so such a load bypasses L1 by construction, on every call,
however weak its ordering.} Relaxed removes ordering, not coherence.

\begin{figure}[h]
\centering
\includegraphics[width=\linewidth]{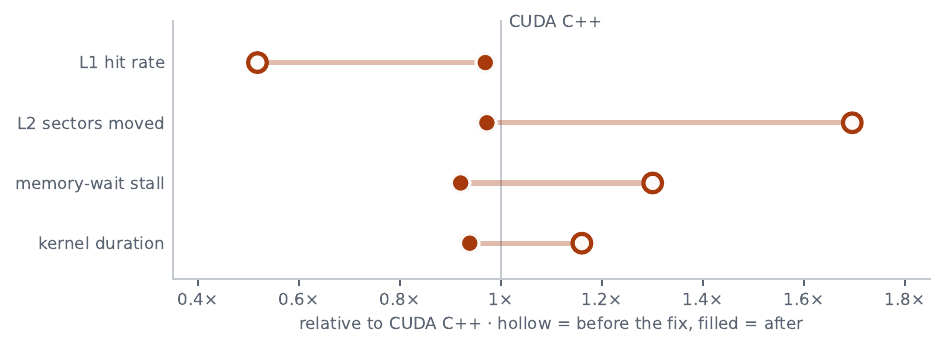}
\caption{What the two-line change moved, each counter normalised to CUDA C++ so
that quantities in different units share one axis. Hollow markers are the Rust
kernel before the fix, filled markers after. The kernel was moving
$1.70\times$ the L2 sectors from identical requests and retaining barely half
as much in L1; afterwards every counter lands on the reference. Note that for
L1 hit rate a value below $1$ is the bad direction, since it is the only
counter here where more is better.}
\label{fig:counters}
\end{figure}

The block index is almost always already published, so the Rust version paid an
uncached access on the common path where the CUDA version pays a cached one and
reserves the uncached access for when it is genuinely waiting. Making both first
reads plain restores every counter to parity (Figure~\ref{fig:counters} and
Table~\ref{tab:counters}) and takes allocate from $1.63\times$ to $1.33\times$
at the baseline cell.

This is correct rather than a relaxation of correctness, and the distinction
matters: the \emph{algorithm} provides the guarantee that the memory model was
being asked for. A stale key costs one extra probe; a stale index merely enters
the wait loop, where an atomic load does observe the publication.

\paragraph{Why we think this generalises.} The construct that is safe, idiomatic,
and type-correct is the expensive one, and nothing in the source, the generated
PTX, the machine code, the instruction counts, or the occupancy reveals it. On a
GPU, ``make this read safe against concurrent writers'' and ``make this read
fast'' are in tension in a way no type system expresses, because coherence and
caching are the same mechanism. Any language that offers a scoped atomic load as
the natural way to read shared state inherits this.

\subsection{Two design experiments that failed}

Both are reported because they bound the space of fixes.

\paragraph{Publishing key and index atomically.} A 128-bit compare-exchange
would publish both words at once and close the window described in
Section~\ref{sec:workload}, removing the wait. The instruction works in
isolation; the algorithm does not. The location must be reserved before the
exchange, so a thread that loses the slot holds one it cannot use, and at kernel
start every thread targeting a block observes the same empty slot before any
exchange lands: 65{,}536 reservations against 1{,}160 real blocks. This rules out
the entire family of fixes that make the location part of the published value.

\paragraph{Deriving the location from the slot.} Setting the block index equal to
the hash slot removes the circularity properly, and is correct. It makes CUDA C++
21--35\% faster and Rust 29--32\% faster, at the cost of sizing the voxel pool to
the table rather than to the block count. It does \emph{not} close the language
gap: $1.64\times$ becomes $1.71\times$ at 320k points and $1.53\times$ becomes
$1.39\times$ at 1.28M.

That last result corrected an earlier claim of ours. Removing the wait from the
Rust arm alone had appeared to recover 75--83\% of the gap, which we had reported
as the wait being most of it. Removing it from both arms shows it is a large
\emph{shared} cost. The asymmetric measurement was the error, and it is the kind
that only a symmetric re-run detects.

\subsection{The residual}

After the fix, Rust's allocate stage spans $1.02$--$1.71\times$ across the
sweep with a median of $1.21\times$, excluding the smallest cell, where the
kernel is short enough that launch effects dominate and the ratio reaches
$3.34\times$. That residual is not attributed. We report it as unattributed
rather than assigning it to the nearest available cause, which over the course of
this project would have been the wrong one eleven times running.

\section{What each language cannot say}
\label{sec:expressiveness}

Section~\ref{sec:attribution} priced what the languages cost. This section
describes what they cannot say, which is what those prices are made of. Every
item below was encountered while implementing the algorithm rather than
constructed to make a point, and we separate genuine expressive limits from one
defect that is merely a bug.

\subsection{Triton: the probe loop}

Listing~\ref{lst:triton} is the Triton probe, against the CUDA C++ of
Listing~\ref{lst:insert}.

\begin{lstlisting}[caption={The same probe in Triton. Both differences from
Listing~\ref{lst:insert} are forced.}, label={lst:triton}]
done = ~active
for p in tl.static_range(0, MAX_PROBE):   # fixed trip count, no early exit
    slot = (start + p) & hash_mask
    eff  = tl.where(done, my_scratch, slot)   # resolved lanes aimed elsewhere
    key  = tl.load(table_i64 + eff * 2)
    ...
    eff_cas = tl.where(empty, slot, my_scratch)
    old = tl.atomic_cas(table_i64 + eff_cas * 2, cmp, want)  # no mask parameter
    done = done | won | hit
\end{lstlisting}

Two things are forced.

\paragraph{There is no per-lane early exit.} Triton programs operate on blocks of
lanes and the compiler owns the mapping from tensor elements to lanes. Allowing
one element to leave a loop would tie control flow to that mapping and forfeit
the compiler's freedom to choose it, which is the freedom Triton's performance on
tiled workloads comes from. The restriction is principled. It is also why the
loop above runs to a bound with a \texttt{done} mask.

The masks are being used outside their design. A \texttt{mask} on
\texttt{tl.load} exists for ragged tile boundaries, where it suppresses a handful
of lanes on one block out of many and its cost is negligible by construction.
Carrying a \texttt{done} mask through a probe loop suppresses most lanes on most
blocks for most iterations, and nothing in the cost model was built to make that
cheap.

\paragraph{\texttt{tl.atomic\_cas} takes no mask, and \texttt{tl.atomic\_add}
does.} This asymmetry looks like drift rather than design: if masking an atomic
were impossible, neither would have it. The consequence is that resolved lanes
must be given somewhere harmless to write, so the programmer acquires a scratch
region, its sizing, and its indexing, none of which appear in the CUDA version.
Section~\ref{sec:attribution} measured the cost of getting that sizing wrong at a
factor of twenty, with the worst setting also destroying the kernel's ability to
use additional hardware.

\subsection{Triton: an unbounded probe is inexpressible at any price}

The bound is a \texttt{tl.static\_range} trip count, so it must be a compile-time
constant and the loop is unrolled. Raising it is not a workaround. Moving from 8
to 32 took roughly twenty minutes to compile against seconds, and produced a
$3.7\times$ larger cubin.

The consequence is a correctness one. A bound small enough to compile is small
enough to lose data, and the sweep now prices that across five load factors
rather than the two we first reported:

\begin{center}
\begin{tabular}{lrrrr}
\toprule
Cell & Load factor & Blocks (CUDA C++) & Blocks (Triton) & Contributions dropped \\
\midrule
lf-sparse    & 0.018 & 1{,}160  & 1{,}160  & 0 \\
base         & 0.035 & 1{,}160  & 1{,}160  & 0 \\
lf-lo        & 0.071 & 1{,}160  & 1{,}159  & 155 \\
lf-mid       & 0.142 & 1{,}160  & 1{,}159  & 155 \\
tartan-p002  & 0.192 & 25{,}230 & 25{,}230 & 22 \\
lf-hi        & 0.283 & 1{,}160  & 1{,}147  & 27{,}943 \\
tartan-dense & 0.437 & 14{,}332 & 14{,}278 & 53{,}614 \\
\bottomrule
\end{tabular}
\end{center}

Every row reproduced identically in all six passes, so the loss is
deterministic rather than a race. Two things in that table were not visible
before. Loss begins at a load factor of $0.071$, not at the $0.283$ we
previously reported as the first failing point; between $0.035$ and $0.071$ the
arm goes from exact to losing a block. And \texttt{tartan-p002} is real depth
data, at a load factor reached by an ordinary trajectory at the pool size every
other real cell uses. We did not have to contrive that cell. It is one of six
RetroOffice trajectories, and it is the one where the volume happens to be
dense enough for a bounded probe to run out.

The failure is also not one thing. At $0.192$ the block set is exact and 22
point contributions are discarded during the update stage, so the surface is
complete and slightly wrong. From $0.283$ upward whole blocks fail to be
allocated and their surface is missing entirely. A consumer would notice the
second and not the first.

\begin{figure}[h]
\centering
\includegraphics[width=0.56\linewidth]{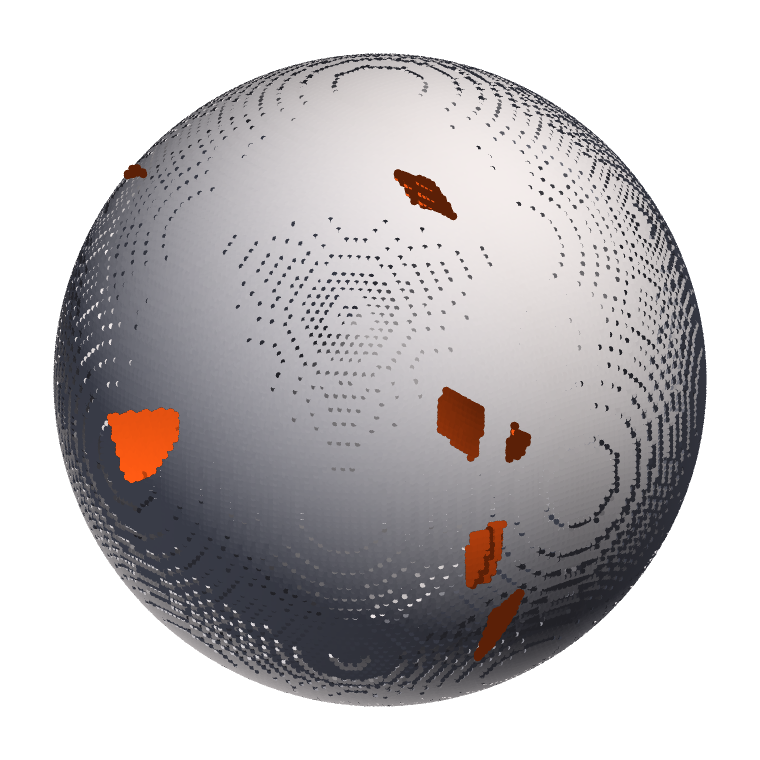}
\caption{What that silence looks like. The surface is the volume CUDA C++ built
at a load factor of $0.283$; marked in colour are the points present there and
absent from the volume Triton built from identical input. The loss is not
scattered, which is the part a count does not convey: it arrives as whole
contiguous patches, because the unit that fails to be allocated is a block of
$8^3$ voxels, and a block that loses its probe takes its entire surface with
it. A reconstruction pipeline consuming this would see holes, not noise.
On real depth the same failure has a different signature. At a load factor of
$0.437$ the \texttt{tartan-dense} cell loses 54 blocks scattered across the
whole room rather than gathered into a few patches, because real occupancy is
irregular where a sphere's is not. That is the harder case to notice: pepper
through a reconstruction reads as sensor noise, and a crater does not.}
\label{fig:loss}
\end{figure}

It was also unreported. The implementation counted blocks lost to pool
exhaustion, as the other arms do, but had no equivalent accounting for probe
exhaustion, because the CUDA and Rust arms cannot exhaust a probe: they scan
until they find a slot. We added the accounting, and note that its absence was
the natural consequence of porting an algorithm whose failure mode does not exist
in the original.

So the honest statement is stronger than a throughput ratio. \emph{Triton cannot
express an unbounded probe. The programmer must pick a bound. A bound small
enough to compile is small enough to lose data at load factors CUDA C++ handles
exactly, and the loss is silent, load-dependent, and absent from sparse
benchmarks.}

\subsection{Triton: diagnostics}

Issuing an atomic on a scalar pointer rather than a tensor of pointers aborts the
compiler with an MLIR assertion, \texttt{only integers and floats have a
bitwidth}, and no source location. A compiler crash without a location is
materially worse than a type error: it reports that something failed, not what
the programmer did. We mention it because it is the kind of cost that never
appears in a performance table and dominates the experience of writing the code.

\subsection{Rust: the safe construct is the expensive one}

Rust's limit is not that something is inexpressible. It is that the natural,
idiomatic, type-correct way to read a location another thread writes carries a
guarantee stronger than the algorithm needs, and that guarantee has a fixed
hardware cost.

A scoped atomic load is coherent at its scope. No NVIDIA L1 is coherent across
streaming multiprocessors. Therefore the access cannot be served from L1, on
every call, regardless of ordering: relaxed removes ordering, not coherence.
Section~\ref{sec:attribution} measured that at 27 points of L1 hit rate and
$1.70\times$ the traffic to L2.

Nothing warns the programmer. The source looks correct, the generated PTX looks
correct, the machine code looks correct, and the instruction counts favour the
slower version. The cost is visible only in where the data is permitted to live.

The corresponding CUDA C++ is a plain load whose safety rests on an argument
about the algorithm rather than on the type system: a stale key costs one extra
probe, and a stale index enters a wait loop. That argument is not checkable by a
compiler, which is exactly why the safer-looking construct is the one a
conscientious programmer reaches for.

\subsection{A defect, which is not the same thing}

We separate the following from the above because it is a bug, and it has been
fixed.

In the version of \texttt{cuda-oxide} used for this work, the scoped atomic load
and store could not be called at all. Both lowered to LLVM IR forms that libNVVM
rejects, so every call failed the backend verifier in the build mode that
produces real kernels, with an error naming an LLVM basic block rather than the
offending line of Rust. Any \texttt{AcqRel} or \texttt{SeqCst} ordering failed
the same way through a rejected LLVM fence. The project's own atomics example
does not build in that mode.

The underlying PTX instructions have existed since sm\_70, so this is a lowering
gap rather than a hardware limit. We contributed a fix that lowers the three
operations to those instructions, along with tests, a regression example, and the
continuous-integration registration whose absence had allowed the defect to
survive: the example that exercises these APIs was only ever run in the build
mode where the problem does not appear. It was reviewed and merged
upstream~\cite{cudaoxidepr}, so the API is callable in current
\texttt{cuda-oxide} and the workaround this paper describes is no longer needed
to reproduce it.

The maintainer amended the patch in two respects, and both are corrections
rather than polish. We had made the assembler's memory clobber conditional,
omitting it for \texttt{Relaxed}; it is unconditional in the merged version,
because without it the compiler may reorder plain accesses to the same address
across the inline assembly and break the single-thread coherence Rust
guarantees regardless of ordering. This does not affect our measurements, which
were taken with the clobber present, but it does mean the version we submitted
would have been wrong in a way our benchmark could not have detected. Sequential
consistency, which we left unsupported, now lowers to a fence fused with an
acquiring load or releasing store.

This correction also revises a claim we made in an earlier draft. We had written
that libNVVM's restrictions force Rust back onto volatile reads and an explicit
fence, and that this weakens the safety argument for Rust on GPUs. That is wrong.
The scoped API exists precisely so callers need not do that; it was simply
unusable. The distinction matters, because a design limitation and a bug support
very different conclusions, and only one of them is an argument about the
language.

\section{Threats to validity}
\label{sec:threats}

\paragraph{One architecture.} Every measurement is on sm\_120, across two GPUs
differing $2.33\times$ in streaming multiprocessor count and roughly $2\times$ in
memory bandwidth. A second architecture was considered and not run.

The magnitudes are therefore Blackwell figures and should be read as such. The
mechanisms are not: Triton's missing per-lane early exit and unmasked
compare-exchange are properties of the language; a \texttt{tl.static\_range}
trip count being a compile-time constant is a property of the language; and a
GPU-scope atomic load bypassing L1 follows from no NVIDIA L1 being coherent
across streaming multiprocessors, which holds on every current architecture. We
claim the ranking and its causes generalise, and that the numbers attached to
them do not. Nothing in the argument rests on a magnitude.

The two-card axis is a partial mitigation rather than a substitute: it varies
machine width at fixed code generation, which rules out the ratios being an
artefact of one machine's balance but says nothing about a different generation.

\paragraph{The sharpest prediction is untested.} As Section~\ref{sec:results}
describes, the load factor reachable in this data structure caps at 0.283, which
is too low to exercise the regime the Triton explanation makes its strongest
prediction about. Testing it requires decoupling table size from pool capacity,
which changes the shared structure and therefore every arm. The prediction is
untested rather than unsupported.

\paragraph{An unattributed residual.} After the fix in
Section~\ref{sec:attribution}, Rust's allocate stage remains $1.02$--$1.69\times$
CUDA C++ with a median of $1.21\times$, and we do not know why. Given that eleven prior hypotheses about this
same gap were each plausible and each wrong, we prefer to leave it open than to
assign it to whichever cause is nearest.

\paragraph{Two environments of real data.} The real-data cells are nine valid
cells from two TartanAir environments: six trajectories of RetroOffice and
three of AmericanDiner. An earlier version of this work used one frame of one
trajectory, and we flagged that as the weakest evidence in the paper. Widening
it changed a conclusion rather than confirming one, which is the argument for
having done it: real depth turns out to sit at the worse end of Triton's range,
not the middle, and one of the six RetroOffice trajectories reaches a load
factor at which the Triton arm silently discards data. Neither was visible from
a single scene. Two indoor environments from one simulator are still not a
claim about real depth data in general, and in particular both are synthetic
renderings with exact ground-truth depth, so they have none of the noise,
dropout or calibration error of a physical sensor.

\paragraph{We wrote all the implementations.} Every arm was written by the same
authors, who had substantially more experience in CUDA C++ than in Triton before
this work. We mitigated this by holding the algorithm, memory layout, hash
function and launch geometry identical across arms, by gating on correctness
before timing, and by attributing every gap to a specific construct rather than
resting on the totals. The Triton results in particular should be read as what a
competent non-expert obtains from the documented API, and an expert may do
better. We note that two of our own Triton and Rust implementations were
materially improved during this work, in both cases by measurement rather than by
insight, which suggests the remaining implementations may not be optimal either.

\paragraph{One workload family.} All cells are TSDF fusion. The properties we
claim to be measuring, data-dependent per-lane work, compare-exchange insertion
and contended scatter, are shared with other GPU hash tables, sparse structure
builders and dynamic work queues, but we have not measured those. The
generalisation in Section~\ref{sec:workload} is an argument, not a result.

\section{Related work}
\label{sec:related}

\paragraph{GPU programming model comparisons.} Performance-portability studies
compare CUDA, HIP, SYCL, Kokkos, RAJA and OpenMP across vendors, typically using
proxy applications drawn from scientific computing~\cite{davis2024portability}.
Those proxies are dense and structured: stencils, particle updates, linear
algebra. Our contribution is orthogonal rather than competing. We hold the
architecture and the algorithm fixed and vary only the source language, and we
choose a workload whose defining property is that the work per thread is not
known at launch. To our knowledge no existing comparison targets that regime,
which is also the regime in which we find the differences to be an order of
magnitude rather than a few percent.

\paragraph{Triton.} Triton~\cite{tillet2019triton} is built around tiles, and its
central design decision is that the programmer describes what data a block needs
while the compiler chooses how that data maps to lanes. Section
\ref{sec:expressiveness} argues that the restrictions we encounter follow from
that decision rather than from oversight, and that the workload we measure sits
as far from its target as a GPU kernel can while remaining a GPU kernel. We
intend the result as a boundary measurement for a tool that is very good inside
its boundary, not as a criticism of the design.

\paragraph{Volumetric fusion.} The truncated signed distance
representation~\cite{curless1996volumetric} and its real-time
use~\cite{newcombe2011kinectfusion} are long established, and the spatial hash
that makes it affordable at scale is due to Nie{\ss}ner et
al.~\cite{niessner2013voxelhashing}. We take the data structure from that line of
work and contribute nothing to it; the reconstruction literature supplies our
workload rather than our question. Open3D~\cite{zhou2018open3d} provides the
third-party implementation used as an external correctness reference, and the
evaluation data are TartanAir V2 sequences, from the dataset introduced
in~\cite{wang2020tartanair}.

\paragraph{Rust on GPUs.} \texttt{cuda-oxide}~\cite{cudaoxide} compiles Rust to
PTX through libNVVM. Published comparisons against CUDA C++ are, as far as we can
determine, absent: the project's own comparison appendix is a placeholder. This
paper is intended partly to fill that gap, and Section~\ref{sec:expressiveness}
reports a defect we found and fixed in the process.

\paragraph{Positioning.} The closest work to ours in spirit is the
performance-portability literature, which asks whether one source can run well
everywhere. We ask a narrower question on harder ground: given one architecture
and one algorithm, what does the choice of source language cost, and can that
cost be attributed to something the language does or does not let you say.

\section{Conclusion}
\label{sec:conclusion}

We implemented one irregular GPU kernel in three languages and measured it on a
matrix of workloads across two GPUs. The result is a split rather than a
leaderboard. On the regular stage the languages differ by a small factor; on the
irregular stage Rust stays close to hand-written CUDA C++ and Triton is more than
an order of magnitude behind. Language choice is nearly free on the work that is
usually benchmarked and expensive on the work that is not, which is an argument
for benchmarking the other work.

Both gaps are attributable, and neither is attributable to code generation
quality in the sense that phrase usually carries. Triton's cost follows from a
probe loop that must run to a compile-time bound and from a compare-exchange that
cannot be masked; the second forces a data structure with no counterpart in the
CUDA implementation, whose bad settings also stop the kernel responding to
additional hardware while still reporting a plausible single-device time. Rust's
cost was invisible to every instruction count we could take, including counts
that favoured Rust, and turned out to be that a GPU-scope atomic load cannot be
served from L1 because coherence and caching are the same mechanism. The
type-correct way to read shared state is the expensive one, and no artefact
short of hardware counters shows it.

If there is a single transferable claim, it is that one. \emph{On a GPU, making a
read safe against concurrent writers and making it fast are in tension that no
type system expresses.} Any language offering a scoped atomic load as the
obvious way to read shared state inherits it, and the alternative is a plain load
whose safety rests on an argument about the algorithm that a compiler cannot
check. That is an uncomfortable place for a language whose appeal is that the
compiler checks things.

We also report the measurement discipline, because three of the errors caught
during this work would have produced a reversed or null conclusion rather than a
noisy one, and two survived until after the wrong number had been written down.
All three were failures of comparability rather than precision: the two sides
were not doing the same work, or were not on the same machine, or were not
summarised by a statistic that survived repetition. None would have been caught
by more repetitions or tighter error bars. We suggest that comparisons of this
kind state how they established those three properties, because in our case each
was something the harness had to verify rather than assume.

Two results we would have preferred not to publish are included because omitting
them would misrepresent how the work went. An earlier version of this paper
reported Triton's allocate gap as $73\times$; correctly configured it is
$17.9\times$, and most of the difference was our own badly-sized workaround
rather than the language. And we previously attributed most of Rust's gap to the
publication wait, on a measurement that removed that wait from one arm while
leaving it in the other. The corrected figures are smaller and the corrected
claims are narrower. They are also the ones that survived a symmetric test.

\paragraph{Future work.} The load-factor regime where probe chains genuinely
lengthen requires decoupling the table size from the pool capacity, and would
test the sharpest prediction the Triton explanation makes. A second architecture
would establish whether the magnitudes hold across a hardware generation; we
expect the mechanisms to and the numbers not to. And the residual
$1.02$--$1.69\times$ on Rust's allocate stage remains open, which we note with
some caution, having been confidently wrong about this particular gap eleven
times already.

\paragraph{Artefacts.} The harness, all five implementations, the raw
measurement CSVs, and the scripts that generate every table in this paper are
available. No number here was transcribed by hand.

\section*{Availability}

The harness, all five implementations, the raw measurement CSVs from every sweep,
and the scripts that generate each table in this paper are available with the
source. Every table is produced from the CSVs by script; no measurement in this
paper was transcribed by hand.

\section*{Disclosure of AI assistance}

This work was carried out with substantial assistance from a large language
model (Claude Opus 5, Anthropic). Specifically, the model wrote the five
implementations, the measurement harness and the analysis scripts, executed the
measurements, and produced the first draft of this manuscript, working
interactively under the author's direction across a single extended session.

The author defined the study, directed the investigation, and made the
decisions that shaped it: which hypotheses to pursue, when to stop measuring and
publish, what to exclude, and how to interpret the results. All measurements were
run on the author's hardware and the raw data are published so that any claim
here can be checked independently of how it was produced. The author has reviewed
the manuscript and is responsible for its content.

We note this workflow is visible in the paper's record of corrections. Several
figures reported in earlier drafts were later found to be wrong and are corrected
in place rather than removed, including two in the conclusion. Those corrections
came from measurement rather than from review, which is an argument for the
methodology in Section~\ref{sec:method} and not for the workflow.

\bibliographystyle{plain}
\bibliography{refs}

\end{document}